\documentclass[sigconf,natbib=false]{acmart}
\usepackage[utf8]{inputenc}
\usepackage{amsmath}
\usepackage{graphicx}
\usepackage{float}
\usepackage{algorithm}
\usepackage{algpseudocode}
\usepackage{multirow}

\copyrightyear{2022}
\acmYear{2022}
\setcopyright{acmlicensed}\acmConference[ICAIF '22]{3rd ACM International Conference on AI in Finance}{November 2--4, 2022}{New York, NY, USA}
\acmBooktitle{3rd ACM International Conference on AI in Finance (ICAIF '22), November 2--4, 2022, New York, NY, USA}
\acmPrice{15.00}
\acmDOI{10.1145/3533271.3561727}
\acmISBN{978-1-4503-9376-8/22/10}

\RequirePackage[
  datamodel=acmdatamodel,
  style=acmnumeric,
  ]{biblatex}
  
\begin{document}

\title{LaundroGraph: Self-Supervised Graph Representation Learning for Anti-Money Laundering}

\author{Mário Cardoso}
\email{mario.cardoso@feedzai.com}
\affiliation{%
  \institution{Feedzai}
  \country{}
}

\author{Pedro Saleiro}
\email{pedro.saleiro@feedzai.com}
\affiliation{%
  \institution{Feedzai}
  \country{}
}

\author{Pedro Bizarro}
\email{pedro.bizarro@feedzai.com}
\affiliation{%
  \institution{Feedzai}
  \country{}
}

\begin{abstract}
    Anti-money laundering (AML) regulations mandate financial institutions to deploy AML systems based on a set of rules that, when triggered, form the basis of a suspicious alert to be assessed by human analysts. Reviewing these cases is a cumbersome and complex task that requires analysts to navigate a large network of financial interactions to validate suspicious movements. Furthermore, these systems have very high false positive rates (estimated to be over 95\%). The scarcity of labels hinders the use of alternative systems based on supervised learning, reducing their applicability in real-world applications.
    In this work we present LaundroGraph, a novel self-supervised graph representation learning approach to encode banking customers and financial transactions into meaningful representations. These representations are used to provide insights to assist the AML reviewing process, such as identifying anomalous movements for a given customer. LaundroGraph represents the underlying network of financial interactions as a customer-transaction bipartite graph and trains a graph neural network on a fully self-supervised link prediction task. We empirically demonstrate that our approach outperforms other strong baselines on self-supervised link prediction using a real-world dataset, improving the best non-graph baseline by $12$ p.p. of AUC. The goal is to increase the efficiency of the reviewing process by supplying these AI-powered insights to the analysts upon review. To the best of our knowledge, this is the first fully self-supervised system within the context of AML detection.
\end{abstract}



\maketitle

\section{Introduction}

\begin{figure*}[!ht]
\centering
\includegraphics[width=\textwidth, height=0.35\textheight]{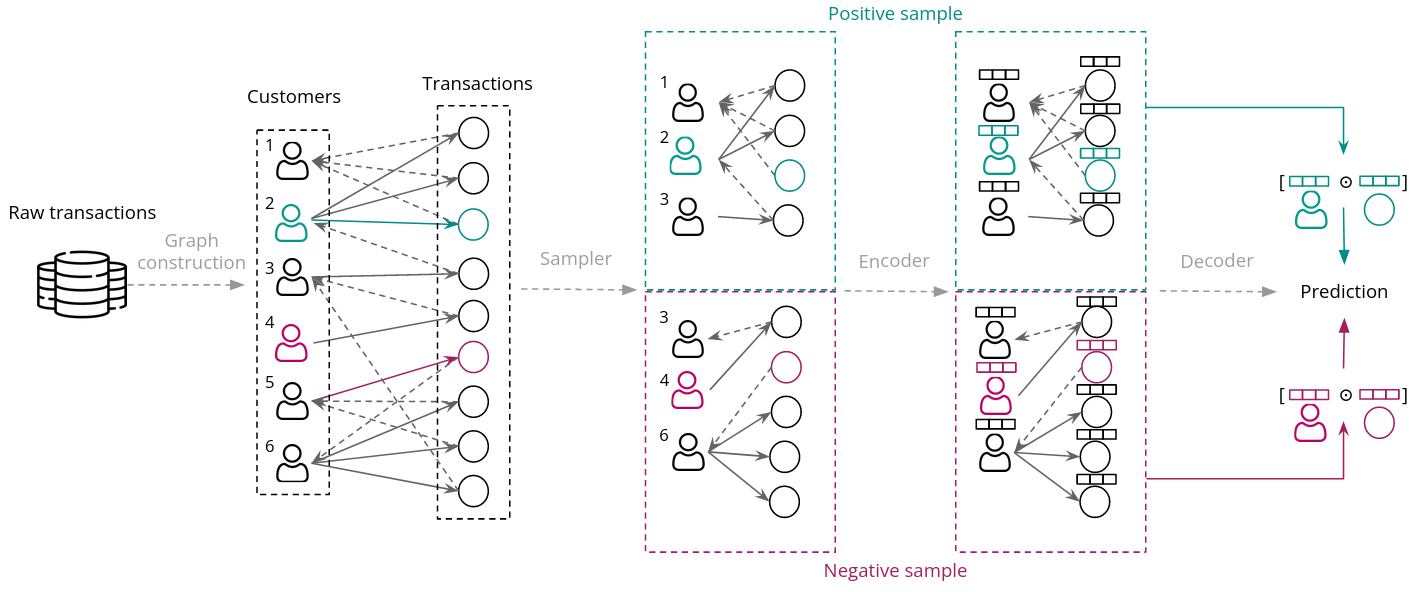}
\caption{Proposed system training overview. Outgoing transactions are represented with filled arrows, and incoming transactions with dashed arrows. First, the bipartite graph is built from a dataset comprised of raw transactions. Then, positive pairs (green) and negative pairs (red) together with their $K$-hop subgraphs ($K=2$ in the figure) are extracted and their embeddings obtained through the encoder. Finally, the decoder uses the aforementioned embeddings to generate the prediction for each sampled edge. \label{fig:overview}}
\end{figure*}

Money laundering is a criminal activity concerned with concealing the origin of funds obtained through illegal means such as terrorism financing, drug trafficking or corruption, appearing legitimate until a thorough analysis is performed. An estimated €1.7 to €4 trillions (2\% - 5\% of global GDP) are estimated to be laundered annually \cite{lannoo2021aml}. To adhere to the AML regulations, financial institutions employ compliance experts that investigate suspicious activities alerted, usually, through a rule-based system. These triggered rules are the starting point of a process that can take several days to complete, culminating in a decision of flagging as suspicious activity or not. When the former is identified, a suspicious activity report must be filed and delivered to a regulatory institution that proceeds with due action. Non-compliance in reporting money laundering can lead financial institutions and their employees to face civil and criminal penalties, such as heavy fines or prison time.

In Anti-Money Laundering (AML) reviewing, analysts investigate alerts centered on an entity (e.g., bank accounts or customers), comprised of a bulk of transactions that triggered one or more rules in order to understand if any suspicious activity was involved. Navigating the network of interactions sprawling from a complex alert and keeping track of the flows of money, often times through entities not directly connected to the one being investigated, is a challenging and cumbersome task. To facilitate this procedure, analysts resort to understanding the data through aggregations of meaningful categories, such as grouping by entities interacted with (known as counterparts) or amounts, as well as relying on their past experience and prior knowledge of the customer under review. Throughout the review process, there is a continuous effort to filter the large bulk of transactions into a smaller set of abnormal interactions that can be used to justify suspicious activity. There are some challenges with the current reviewing process, namely: 1) New analysts lack the context more experienced analysts might have, requiring an additional effort to familiarize themselves with re-occurring customers. Similarly, additional effort is required to contextualize new customers entering the system; 2) It is challenging to navigate the bulk of transactions and decide which movements are particularly suspicious, and resorting to a macro-view of the interactions can lead to missing the fine-grained details of each transaction.

To mitigate the aforementioned challenges, in this work we present \textbf{LaundroGraph}, a novel fully self-supervised approach leveraging Graph Neural Networks (GNNs) to encode representations of customers and transactions within the context of AML reviewing. We propose to represent the network of financial interactions as a directed bipartite customer-transaction graph \footnote{Other networks were considered but this was simultaneously the best performing and most flexible approach}, with the GNN trained through a link prediction task between pairs of customer and transaction nodes, essentially corresponding to an anomaly prediction task. As a result, anomalous movements within the context of each customer can be automatically identified and shown to the analyst upon review, providing a starting point of potentially suspicious movements and alleviating the effort required to filter the bulk of transactions. Furthermore, the derived representations can be used as building blocks for additional insights to support the reviewing process, such as clustering the per-customer transactions, and comparing how the behavior of a customer evolves over time. The former can be a useful approach to group the information shown to the analyst beyond simple aggregations, and the latter can quickly provide context surrounding a customer under review. Unlike most existing works in the graph self-supervised literature landscape, in this work self-supervision is both the starting point and the end goal, as there are no anomaly labels or supervised downstream tasks. The objective is for this system to be integrated within a broader system for AML reviewing that handles the necessary workload of assessment creation. Within this system, these insights will be digested and provided in an easy-to-understand manner through tailor-made visualizations for AML as soon as the investigation starts. These visualizations are beyond the scope of this work and they will not be described.

In summary, this work's main contributions are:
\begin{itemize}
  \item A novel fully self-supervised approach to derive representations of customers and financial transactions useful for a variety of insights to support the AML reviewing process.
  \item A new way to represent the network of financial interactions as a customer-transaction bipartite graph.
  \item Validation of our method on a real-world banking dataset in the self-supervised task of link prediction, achieving an improvement of $12$ p.p. of AUC compared to using only the raw features.
\end{itemize}

\section{Related Work}

Most of the approaches to detect AML used by financial institutions are based on a set of rules aligned with regulations. Machine learning methods for AML are becoming more popular, and can broadly be separated into supervised and unsupervised approaches, with the latter being more common due to the lack of available labels. When labels are available, several works have compared the performance of different classifiers and training strategies in predicting money laundering. Examples include benchmarking several popular classifiers and sampling schemes \cite{zhang2019amlsampling}, comparing the performance of an XGBoost classifier when trained exclusively with alerted events or with all events \cite{martin2020aml}, and comparing the performance of an SVM classifier under different hyperparameter configurations \cite{keyan2011svm}. 

Unsupervised approaches typically apply an anomaly detection algorithm by comparing events with the expected behavior through deviation metrics. Definitions of expected behaviour include clusters of transactions by the same customer \cite{larik2011clustering}, the nearest large cluster \cite{gao2009lof}, or the k-nearest neighbors \cite{luna2018shell}.
To handle the lack of real-world data, several approaches have proposed to generate synthetic data, either generating entire datasets \cite{luna2018shell, luo2014amlsup, weber2018aml}, or just patterns of suspicious behavior \cite{gao2009lof, wang2009mst}.

The majority of works using machine learning for AML rely entirely on feature sets that characterize individual events or entities. This naturally disregards the underlying contextual information that is crucial in identifying suspicious behavior. Recent approaches have sought to incorporate such information to improve performance by leveraging the underlying graph of interactions. This is typically done either by explicitly calculating additional features based on the graph \cite{oliveira2021gw, naser2021aml}, or implicitly through node embedding approaches \cite{weber2018aml, weber2019bitcoin, lo2022inspectionL, hu2019node2vecaml}. \citeauthor*{oliveira2021gw} \cite{oliveira2021gw} derive a set of new features based on the structure of the graph by collecting a variety of metrics based on random walks. This work is afterwards extended \cite{naser2021aml} through a triage model that sits downstream of the triggered rules to reduce the number of false positives. This triage model is comprised of a classifier that operates on an extended feature set to predict the risk of an alert. 

Instead of explicitly calculating metrics based on graph neighborhoods, an alternative approach is to automatically derive representations that exploit the underlying structure according to some objective. Both approaches are compared by \citeauthor*{hu2019node2vecaml} \cite{hu2019node2vecaml}, and representations derived implicitly through node2vec \cite{grover2016node2vec} and deepwalk \cite{perozzi2014deepwalk} achieved better downstream classification results than using curated features. Similarly, Weber at al. \cite{weber2019bitcoin} compare a variety of supervised machine learning models, including the popular graph convolutional network (GCN) \cite{kipf2016gcn}, in predicting if a transaction is illicit. Interestingly, the authors found that the GCN performs worse than a random forest, which they justify by concluding that the input features are quite informative, as they already contain a plethora of curated features that characterize a node's neighborhood. Regardless, they note that extending the feature set with the embeddings derived by the GCN model improves upon all the results, further supporting that argument that implicitly derived representations contain additional, meaningful information. Following this work, \citeauthor*{lo2022inspectionL} \cite{lo2022inspectionL} further improve upon the results by leveraging the popular Deep Graph Infomax (DGI) \cite{velickovic201dgi} self-supervised objective to extract additional input features. Contrary to our work, in this study self-supervision is used as a stepping stone to improve the supervised task results, as opposed to being the focal point. Regardless, to the best of our knowledge, this is the first and only work combining self-supervised GNNs and AML detection.

\section{Method}

In this section, we start by describing the proposed graph and the construction procedure from the raw dataset. Then, the model architecture is detailed, followed by an overview of the self-supervised objective and training setup.

\subsection{Customer-Transaction bipartite graph}
Translating the information about financial interactions into a graph is a crucial design choice. Motivated by the challenges listed in Section 1, the chosen graph representation should: 1) Maintain the fine-grained nature of the interactions and flow of money; 2) Incorporate new transactions as they enter the system; and 3) Support information at both the customer and transaction level. Given these requirements, we propose a directed bipartite graph comprised of customer and transaction nodes, created through raw data of past transactions performed within a fixed snapshot of time. This graph dictates the representation of behaviour of customers that will be learned, which is used as a reference point to score new transactions entering the system. After enough new data is accumulated, the model can be re-trained on a new graph to capture new behavioural patterns. The choice of a bipartite graph as opposed to a homogeneous multigraph is motivated by two main factors: 1) It trivially allows for the learning of separate latent embedding spaces specific to each node type, which can be used directly or as building blocks to downstream tasks at the level of each node type; and 2) It provides the flexibility to include additional node types that may be relevant in the future, such as merchant nodes or card transaction nodes, with specific properties and features. To further illustrate the first point, in Sections \ref{sec:transaction_umap} and \ref{sec:customer_umap} we study the obtained representations at the transaction and customer level, respectively, and showcase different insights that can be extracted from them.

More formally, we consider a directed bipartite graph $G = (V,E)$, with $V = C \cup T$ denoting the set of customer ($C$) and transaction ($T$) nodes, and $E = I \cup O$ denoting the set of edges between them, where $O$ represents outgoing transactions of the form $C \rightarrow T$, and $I$ represents incoming transactions of the form $T \rightarrow C$. Each node type is associated with a feature vector $f_c \in R^{d_c}$ and $f_t \in R^{d_t}$, respectively representing the customer and transaction node feature vectors. Customer features, which we refer to as profiles, characterize the customers' transactional behaviour within time-windows of different granularities, plus other relevant attributes about the customer, while transaction features contain information about the transaction itself. Customer nodes are connected to all transactions in which they are involved, and transaction nodes are connected to their source and destination customer. As such, each customer has as many edges as transactions performed in that time period and each transactions has, at most, two edges: one incoming and one outgoing. A simplified illustration of this graph can be visualized in Figure \ref{fig:overview}.

\subsection{Self-Supervised anomaly detection}
\subsubsection{Preliminaries}
The objective is to jointly learn an encoder $\mathcal{E} (\mathbf{X},\mathbf{A}) \rightarrow \mathbb{R}^{N_c \times d'_c} \times \mathbb{R}^{N_t \times d'_t}$ and a decoder $\mathcal{D} (\mathbf{z}_c, \mathbf{z}_t) \rightarrow \mathbb{R}^1$. The encoder receives a node feature matrix $\mathbf{X}: \mathbb{R}^{N_c \times d_c} \times \mathbb{R}^{N_t \times d_t}$ and an adjacency matrix $\mathbf{A}: \mathbb{R}^{N_c \times N_t} \times \mathbb{R}^{N_t \times N_c}$ and produces a set of embeddings $\mathbf{Z} = [\mathbf{z}_c^{i}, \mathbf{z}_t^{j}], \forall i \in \{0, ..., N_c\}, j \in \{0, ..., N_t\}$, with each embedding $\mathbf{z}_c^i \in \mathbb{R}^{d'_c}$ and $\mathbf{z}_t^j \in \mathbb{R}^{d'_t}$ denoting the representations for each customer node $i$ and transaction node $j$, respectively. The decoder receives a pair of customer-transaction embeddings $(\mathbf{z_c}, \mathbf{z_t})$, and outputs the likelihood of that transaction existing for that customer.

\subsubsection{Model overview}

We use an encoder comprised of several layers of graph convolutional operators. These operators compute representations by repeatedly sending messages along the edges of a node's local neighbourhood, which are afterwards aggregated and combined with the source node's information. A consequence of this message passing system is that the representations calculated for each node take into account the context surrounding it, a property that is crucial in AML scenarios. The receptive field of each node is defined by the number of layers of the GNN. In other words, the more layers there are, the farther away the neighbours that affect the central node can be. In our experiments, we use the graph attention convolution operator (GAT) \cite{velickovic2018}, defined as follows:
\begin{equation}
    \mathbf{z}^{\prime}_i = \bigg\Vert_{k=1}^K \mathrm{ReLU} \left( \alpha^k_{i,i} \mathbf{W}^k \mathbf{z}_i + \sum_{j \in N_{(i)}} \alpha^k_{i,j} \mathbf{W}^k \mathbf{z}_j \right)
\end{equation}

\begin{equation}
    \alpha_{i,j} = \dfrac{exp(LeakyReLU(\mathbf{a}^T [\mathbf{W} \mathbf{z}_i \Vert \mathbf{W} \mathbf{z}_j])}{\sum_{k \in N_{(i)}} exp(LeakyReLU(\mathbf{a}^T [\mathbf{W} \mathbf{z}_i \Vert \mathbf{W} \mathbf{z}_k])}
\end{equation}

In the previous equations, $N_{(i)}$ denotes neighbors of node $i$, $\Vert$ denotes concatenation, with $\Vert_{k=1}^K$ denoting concatenation over $K$ attention heads, $\alpha_{i,j}$ denotes an attention coefficient between nodes $i$ and $j$, and $\mathbf{W}$ and $\mathbf{a}$ denote learnable parameters. Note that since we have a bipartite graph, nodes $i$ and $j$ will be of different types (i.e., if node $i$ is a customer node, then node $j$ will be a transaction node, and vice-versa), with a different set of learnable parameters per each node and edge type. We hypothesize that the additional expressiveness provided by the attention mechanisms is beneficial in this scenario, particularly in situations where the transaction to classify is similar to an existing interaction, allowing the model to assign a higher attention coefficient to that interaction.

Let $\odot$ denote the Hadamard product and $\sigma$ the sigmoid non-linearity. The decoder is comprised of a simple feed-forward, and the prediction for an edge with customer node $c$ and transaction node $t$ is defined as follows:

\begin{equation}
    \hat{y}_{c,t} = \sigma \left( \mathbf{W} [\mathbf{z}_c \odot \mathbf{z}_t] \right)
\end{equation}

Given this prediction, the anomaly score is defined as $1 - \hat{y}_{c,t}$. Note that we consider a single decoder that predicts both incoming and outgoing transactions. The entire forward propagation procedure is detailed in Algorithm \ref{alg:algorithm}, considering a mini-batch scenario.

\subsubsection{Training objective}
As previously mentioned, the objective is to identify anomalous transactions within the context of a customer's usual behavior. This usual behaviour is dictated by the input graph $G$, and is leveraged by the decoder to classify new transactions entering the system. Since labels are not available, we resort to self-supervision. Typical self-supervised approaches with graphs use the graph structure itself as a means of deriving labels. This commonly translates to sampling positive and negative examples, together with a loss function that promotes the representations of positive/negative samples to be similar/dissimilar, respectively. Concrete examples of this framework are the popular graph auto-encoders (GAE) \cite{kipf2016autoencoder} that seek to reconstruct the adjacency matrix and the subsequent extension to objectives based on random walks \cite{grover2016node2vec, hamilton2017}. Here we follow a similar approach to the GAE, where the network is tasked with predicting the likelihood of an edge existing between the entities sent as input. Positive examples are defined as customer-transaction edges that exist in the graph and negative examples are obtained through the sampling function $S$ in Algorithm \ref{alg:algorithm}, which randomly samples customer and transaction nodes to create non-edges \footnote{An important direction of future work would be exploring alternatives to uniform negative sampling}. In either case, the edges corresponding to the direction being predicted are severed. Given a positive example $(c, t)$, and $M$ sampled negative examples $(\Tilde{c}, \Tilde{t})$ from a negative sampling distribution, the encoder and decoder are jointly trained through a standard binary cross-entropy (BCE), defined as follows:

\begin{equation} \label{eq:4}
    \mathcal{L}(c,t) = - log (\hat{y}_{c,t}) - M \cdot log(1 - \hat{y}_{{\Tilde{c}, \Tilde{t}}})
\end{equation}

Negative examples are only used for training the model. During production, all transactions entering the system are positive examples for which we already know the entities involved. To obtain the corresponding anomaly scores, we follow the same procedure detailed above: the directed edge being predicted is severed, followed by using the encoder to obtain the transaction embedding. This embedding is then used by the decoder together with the previously obtained customer embedding (i.e., the customer's "expected" behavior) to calculate the anomaly score, as described in Section 3.2.

\begin{algorithm}
\caption{LaundroGraph forward propagation algorithm}\label{alg:algorithm}
\begin{flushleft}
        \textbf{Input:} Graph $G$; number of layers $L$; neighborhood sampler $\mathcal{N}$; mini-batch size $B$; edge sampling function $\mathcal{S}$; edge direction $D$
\end{flushleft}
\begin{algorithmic}
\State $E_{p} : {(c_1, t_1),..., (c_B, t_B)} \gets $ select $B$ edges from $G$ in direction $D$
\State $E_{n} : {(\Tilde{c}_1, \Tilde{t}_1), ..., (\Tilde{c}_B, \Tilde{t}_B)} \gets \mathcal{S}(G)$ \Comment{Sample random $c$ and $t$ as non-edges}
\State $E \gets E_{p} \cup E_{n}$
\If{$D ==$ outgoing}
\State $G \gets G \setminus (c\rightarrow t), \forall t \in E$ \Comment{Delete real outgoing edges}
\Else
\State $G \gets G \setminus (t\rightarrow c), \forall t \in E$ \Comment{Delete real incoming edges}
\EndIf
\State $\mathbf{z^0_{c_i}}, \mathbf{z}^0_{t_i} \gets \mathbf{f}_{c_i}, \mathbf{f}_{t_i}, \forall (c_i,t_i) \in (N_{(c)}\cup c, N_{(t)} \cup t), \forall (c,t) \in E$ \Comment{Input to the first layer is the raw features of all required nodes}
\For{$l \in {1,...L}$}
\For{$(c, t) \in E$}
\State $\mathbf{z}^l_c \gets \mathrm{Convolve}(\{\mathbf{z}^{l-1}_{c_i}, \forall c_i \in \mathcal{N}_{(c)} \cup c\})$ \Comment{Encode nodes}
\State $\mathbf{z}^l_t \gets \mathrm{Convolve}(\{\mathbf{z}^{l-1}_{t_i}, \forall t_i \in \mathcal{N}_{(t)} \cup t\})$ \Comment{Encode nodes}
\EndFor
\EndFor
\State $\hat{y}_{c,t} \gets \sigma \left( \mathbf{W} [\mathbf{z}^L_c \odot \mathbf{z}^L_t] \right), \forall (c,t) \in E $ \Comment{Decoder edge prediction}
\end{algorithmic}
\end{algorithm}

\section{Experiments}
In this section, we start by describing the real-world dataset used in our evaluation together with the baselines considered (Section 4.1). Then, we report the classification results on the fully self-supervised task of link prediction (Section 4.2), followed by a visualization of the transaction embeddings (Section 4.3) and customer embeddings over different time-windows (Section 4.4).

\subsection{Experimental setup}
\subsubsection{Dataset}
We use a real-world banking dataset in our experiments whose identity we cannot disclose for privacy reasons. The dataset consists of approximately one year of bank transfers, and we use 6 months of data to create the graph. The customer profiles are calculated based on all the transactions prior to the start of the snapshot used for training, whereas the transaction features contain information about the transaction itself. The features reflect information that is used in a typical rule-based system deployed by financial institutions. In total, there are $66$ customer features, including daily, weekly and monthly aggregates of past behavior, plus other relevant attributes (e.g., country, risk rating and max/min amounts). The transaction nodes contain $12$ features reflecting the properties of that interaction, such as the amount, countries of the banks involved, information regarding the time-stamp, among others. The resulting graph contains $3.2$M customer nodes and $17.7$M transaction nodes, with an average in/out node degree of $5.23/4.94$ for customer nodes, and a total of $17.1$M incoming edges and $16.2$M outgoing edges. During training, we reserve $30\%$ of the edges for supervision and $20\%$ for validation, with the remaining edges used for message passing. The following month of transactions is used as test data, and we report the results on a subset of $100K$ customers with a total of around $514K$ transactions. For each edge, we randomly sample one non-edge as a negative example (i.e., $M = 1$ in Equation \ref{eq:4}). To fit the graph in GPU memory, we use a neighborhood sampling procedure \cite{hamilton2017}, sampling $32$ neighbors in each direction at each layer. The validation set is used to tune model hyperparameters with the early stopping procedure.

\subsubsection{Baselines}
Regarding baselines, we experiment with popular baselines of an MLP and LightGBM that rely on the same feature information as the graph, but disregard the added structural information. In other words, these baselines are tasked with predicting the existence of an edge, given only the raw features of the source customer, destination customer and transaction. Note that the LightGBM is consistently the best algorithm in this use case (i.e., fraud or money-laundering detection with tabular data). For these baselines, we create a dataset with the same number of positive and negative samples by appending the negative samples to the dataset containing all the positive transactions. Negative examples are created by randomly sampling a source customer, destination customer and transaction.

Besides the raw features baselines, we experiment with another popular self-supervised GNN objective, namely the Deep Graph Infomax (DGI) \cite{velickovic201dgi} objective. In this scenario, the encoder is trained to generate node embeddings that summarize meaningful information from the graph and that are agnostic to the downstream supervised task. This is done by maximizing the mutual information between nodes and the graph they belong to, such that a discriminator can distinguish between the real graph and a corrupted one. Similar to the original paper, we define the corrupted graph through a random node re-shuffling. Given that the original DGI was proposed for homogeneous graph, we naively extend it to our scenario by independently applying the DGI objective for each node type. For each type, the real and corrupted graphs are defined considering only nodes of the same type. After the encoder has been trained, the produced embeddings are used to train the decoder on the link prediction downstream task, using an architecture identical to the one described in Section 3.2. During inference, as is done with the remaining variants, the customer embeddings received by the decoder are the ones derived during the training period. While it is also reasonable to update customer embeddings after each new transaction, we leave this as future work. 

Finally, we also consider different variants of the proposed architecture, namely by replacing the GAT \cite{velickovic2018} operator with two other popular convolutional operators: the GraphSAGE operator \cite{hamilton2017} and the GIN operator \cite{xu2018gin}.

\begin{table}[t]
\begin{tabular}{lcc}
\toprule
Method & AUC & AP \\
\midrule
$\mathrm{MLP}$ & $77.26$ & $82.45$ \\
$\mathrm{LightGBM}$ & $82.58$ & $89.02$\\
\midrule
$\mathrm{DGI}$ & $85.87$ & $84.06$\\
\midrule
$\mathrm{LaundroGraph}_{SAGE}$ & $89.97$ & $93.17$\\
$\mathrm{LaundroGraph}_{GIN}$ & $90.24$ & $93.82$\\
$\mathrm{LaundroGraph}_{GAT}$ & $\mathbf{94.83}$ & $\mathbf{95.22}$\\
\bottomrule
\end{tabular}
\caption{ROC AUC and average precision (AP) results on the test data for all methods under consideration. Best values represented in bold.}
\label{tab:results}
\end{table}

\begin{figure}[t]
\centering
\includegraphics[width=\linewidth, height=0.35\textheight]{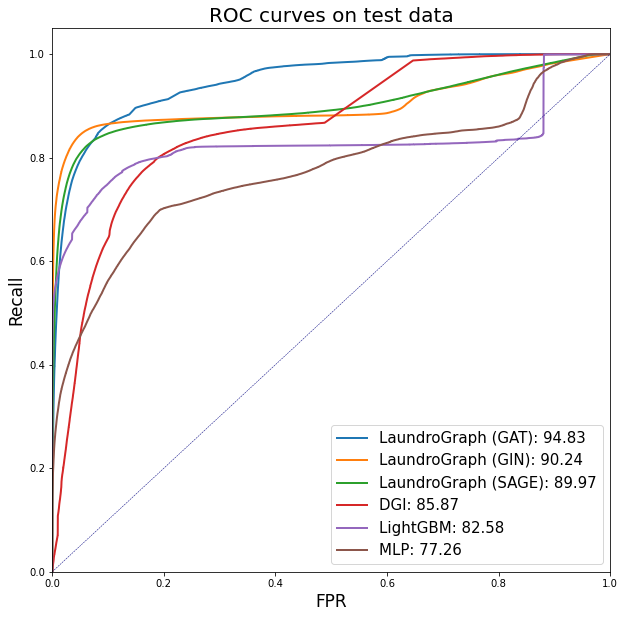}
\caption{ROC curves and corresponding AUCs for all models considered. \label{fig:rocs}}
\end{figure}

\subsection{Classification results}

\subsubsection{Hyperparameters}

\begin{figure*}[!htb]
\centering
\includegraphics[width=\linewidth, height=0.35\textheight]{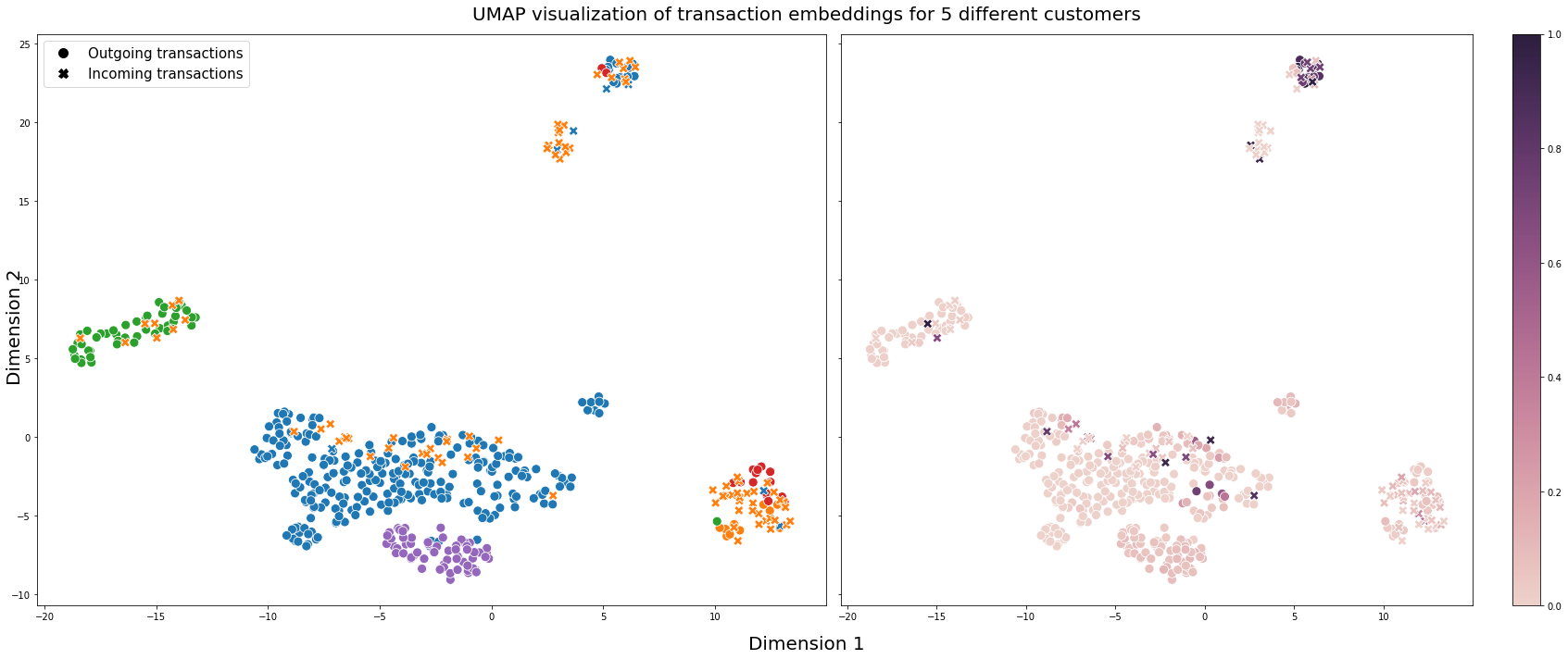}
\caption{UMAP visualization of the transaction embeddings produced by $\mathrm{LaundroGraph}_{GAT}$ for 5 randomly sampled customers. On the left plot, outgoing transactions are represented with a circle marker, and incoming transactions are represented with an X marker. Colors represent the different customers. On the right plot, transactions are colored according to their anomaly score, with darker colors denoting a higher anomaly score. \label{fig:umap}}
\end{figure*}

Hyperparameters are selected through the Tree-structured Parzen Estimator (TPE) \cite{bergstra2011tpe} algorithm using the validation loss as the metric of success. For each model, $20$ different hyperparameter configurations are trained. The hyperparameters were finally set as $5$ layers with dimensions $[128,64,32,16,1]$ and a dropout \cite{srivastava2014dropout} of $0.1$ for the MLP baseline, and $400$ maximum leaves with $150$ minimum samples per leaf for the LightGBM baseline. 

Regarding the graph-based models, the $GAT$ variant has $3$ layers with a hidden size of $32$ and $4$ attention heads. For the SAGE variant, we use the mean aggregator variant \cite{hamilton2017} with $3$ layers with dimension of $256$, together with a skip connection on the source node. For the GIN variant, we use the GIN-0 variant \cite{xu2018gin} comprised of a simple 2-layer MLP per GNN layer. For the DGI baseline, we use the GAT variant described above as the encoder. In all aforementioned neural network baselines, the $\mathrm{ReLU}$ activation function together with batch normalization \cite{ioffe2015bn} is applied in every hidden layer. For training, we use the Adam optimizer \cite{kingma2014adam} with a learning rate of $0.001$ for the GNN-based models and $0.01$ for the MLP baseline. The learning rate for the LightGBM baseline is also set as $0.01$. In all aforementioned baselines, early stopping is applied with a patience threshold of $6$, i.e., we stop training if the validation loss does not improve after $6$ epochs.

Table \ref{tab:results} reports the ROC area under the curve (AUC) and average precision (AP) results on the test data. We note that the raw features are already quite informative, as can be seen by the competitive results achieved with the MLP and LightGBM baselines. Regardless, all graph-based baselines achieved superior performance, showcasing the importance of leveraging the structural information provided by the underlying graph. We further observe that jointly training the encoder and decoder directly on the link prediction task consistently yields better results than training the encoder with the DGI objective, resulting in a difference of $8.95$ and $11.86$ p.p. of AUC and AP, respectively, compared with $\mathrm{LaundroGraph}_{GAT}$. Nevertheless, the strong results achieved by the DGI objective showcase the method's ability to create informative node representations decoupled of any specific task. The GAT variant of the proposed model ($\mathrm{LaundroGraph}_{GAT}$) achieves the overall best results, as the additional expressiveness provided by the attention mechanisms seems to be beneficial in this scenario. The GraphSAGE variant ($\mathrm{LaundroGraph}_{SAGE}$) achieves the worst results of all three convolutional operator variants, which we hypothesize to be due to the lack of expressiveness compared to the remaining three variants, and the dependence on the underlying homophilous nature of the graph.

Figure \ref{fig:rocs} shows the ROC curves of all the methods reported in Table \ref{tab:results}. The ROC curves show the trade-off between recall and specificity. Moreover, the area under the curve (AUC) can be seen as a measure of separability, representing how much a model is capable of distinguishing between classes. From observing Figure \ref{fig:rocs}, we verify that all graph-based models consistently outperform the baselines relying exclusively on raw features. In particular, for very low false positive rates (FPRs), all graph-based variants trained directly on the link prediction task already achieve a recall of $>80\%$, whereas the MLP and DGI baselines achieve a recall of $40\%$ or below, with the LightGBM baseline being a middle-ground between them at $\sim 60\%$ recall. As the FPR increases, the DGI baseline approaches the performance of the remaining graph-based baselines, while the MLP and LightGBM baselines continue to achieve consistently inferior results.

\subsection{Transaction visualization}
\label{sec:transaction_umap}

\begin{figure*}[!htb]
\centering
\includegraphics[width=\linewidth, height=0.3\textheight]{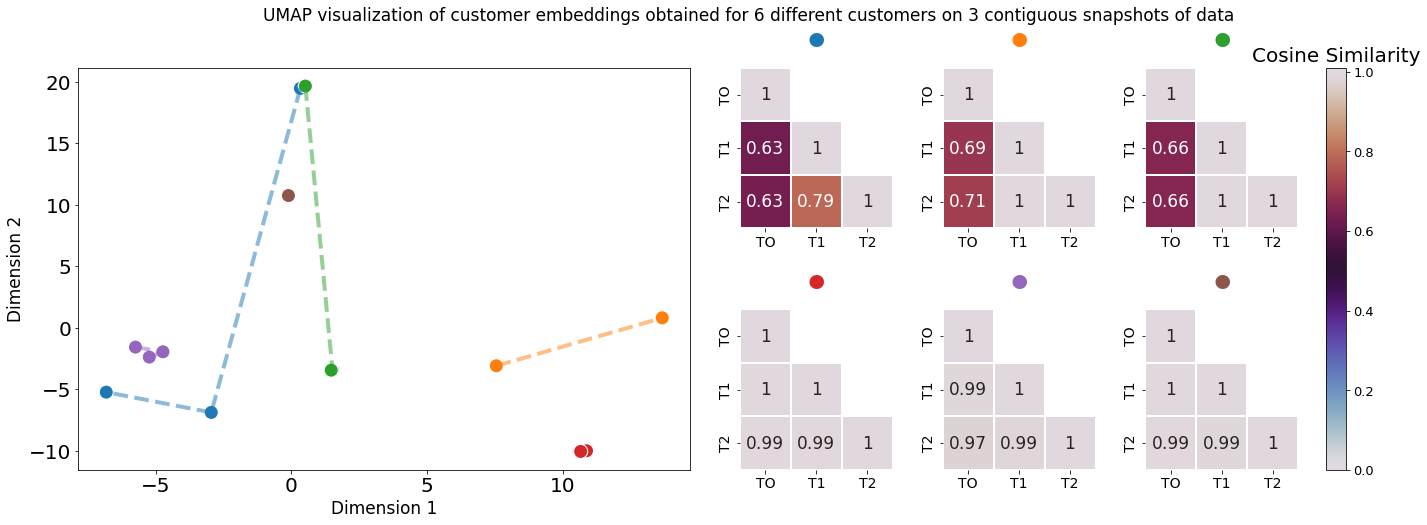}
\caption{UMAP visualization of the customer embeddings produced by $\mathrm{LaundroGraph}_{GAT}$ (left), together with corresponding cosine similarity heatmaps (right), for 6 sampled customers across 3 snapshots of data. Colors represent the different customers. On the left plot, the UMAP embeddings are shown, with each customer providing 3 points, one for each snapshot (18 points in total), connected through a dashed line of the same color. On the right plot, the cosine similarities on the original embedding space are shown, for each customer and snapshot. \label{fig:customers_umap}}
\end{figure*}

Figure \ref{fig:umap} shows a plot of the UMAP \cite{leland2018umap} embeddings for the transactions of 5 different and randomly sampled customers with more than 10 transactions. The marker represents the direction of each transaction, with "o" representing outgoing transactions, and "x" representing incoming transactions. On the left side of the figure transactions are colored according to their customer, and on the right side transactions are colored according to their anomaly score.

From the left side we can observe transactions are naturally clustered according to their customer, and that there are multiple clusters of activity for each customer. We can also observe some level of separability between customers. It is expected for a customer to have several clusters of activity representing the different types of counterparts interacted with, as well as some intracluster variability representing the properties of each transaction. To illustrate this point, we note that during the test period, for the green customer, all outgoing transactions except one were received by the same counterpart, resulting in the left-most green cluster. The remaining outgoing transaction can be seen farther away, near the right-most cluster. At first glance, we may expect this transaction to be anomalous, however, we reiterate the importance of the history observed during the training period, as similar interactions between those two entities occurred frequently. Another interesting scenario is the purple customer. In this case, the cluster represents interactions with several different counterparts whose behaviour is very similar. More specifically, almost all counterparts only received transactions from the purple customer during the training period. From the right side of the figure we can observe that, generally, transactions farther away from their respective non-anomalous clusters (i.e., the "expected" behavior) usually have a higher anomaly score. This can be observed, for example, with the anomalous cluster at the top, and with the scattered incoming transactions from the orange customer.

As detailed in Section 1, aggregating the transactions for a customer under review according to meaningful categories is a key component of the AML investigation process. Aggregating, on-demand, the transactions shown to the analyst according to these clusters manifesting in the latent embedding space goes beyond simple aggregation schemes, grouping the different transactions according to their contextual information and potentially highlighting clusters of normal/anomalous activity.

\subsection{Customer visualization}
\label{sec:customer_umap}

Figure \ref{fig:customers_umap} shows a plot of the UMAP embeddings for $6$ different customers, across $3$ snapshots using a rolling time window, together with corresponding pairwise cosine similarity heatmaps calculated on the original latent embedding space. Each snapshot describes a graph comprised of $6$ months worth of transactions, with each subsequent snapshot sliding the window $1$ month into the future. Comparing the embeddings produced for the same customer across the different windows of time can be seen as a measure of behavior divergence. For the sake of visualization for this example, we consider only customers that have new activity in the differing time periods. Furthermore, since the vast majority of customers retain similar embeddings, we sample half of the customers from the pool of customers with one value of cosine similarity below $0.8$, and the other half from the remaining customers, corresponding, respectively, to the top and bottom half of the heatmaps shown in the right side of the figure.

From the figure we can observe instances of stable and diverging behavior, with divergences observed through shifts in the embedding space, visualized through the associated dashed lines on the left side of the figure, and through darker cells on the right side of the figure. For the customers with stable behavior (i.e., customers with very high cosine similarity values), we note that, in general, the corresponding subgraphs sprawling from the interactions remains largely similar across snapshots. In other words, the new transaction nodes introduced connect to either existing customer nodes, or introduce a new customer node with a neighborhood similar to existing nodes at the corresponding depth. For customers displaying a divergence in behavior, the opposite is observed. Specifically, we note that a common reason for divergence of representations is due to a new type (i.e., incoming or outgoing) of transaction being performed for the first time. This is the case for orange customer, for example. Another observed reason for divergence, exemplified through the blue and green customer, is associated with the counterparts interacted with and the structure of their neighborhoods. As previously detailed, a consequence of the message passing mechanism is that each message contains information about the sender's neighborhood. As such, even if the number and type of transactions performed remains the same across snapshots, a customer can obtain different representations if the received messages describe very different neighborhoods (e.g., due to interacting with new counterparts or if the existing counterparts shift in behavior). This is alleviated for high centrality nodes, as the contribution of each message on the final representation is diminished. In other words, the more we know about a customer's transactional behavior, the more stable their representations will be.

Note that, for this example, the representations used are the ones derived by the last layer of the $\mathrm{LaundroGraph}_{GAT}$ model (i.e., the third layer). By using the representations at different depths of the network, different information can be prioritized, highlighting different types of behavior at the cost of potentially more volatile representations. For example, using the representations of the first layer would provide behavior divergence measures that reflect exclusively the source customer's transactions. Using the representations provided by the second layer additionally considers the counterparts interacted with. In this example, using three layers means that the counterpart's transactions also have an impact on the source representation. Doing so results in more stable representations, where interacting with new entities can lead to similar embeddings if these entities are similar to ones already interacted with in the past. Conversely, if the counterpart's transactional behavior changes drastically between periods of time, then the source embedding will also reflect that, giving an illusion of behavior divergence, as exemplified through the blue customer.

This divergence information can be shown to the analyst to accelerate the contextualization of the customer, providing a continuous macro-view of their behavior that can be used to compare with past decisions. For example, if a customer has had several false positives in the past, and their representation for the current assessment does not diverge drastically from those periods, then it is expected that the current assessment will also be a false positive, introducing a prior on the analyst before any transaction is investigated. 

\section{Conclusion}
In this work, we introduced LaundroGraph, a fully self-supervised approach to support the AML reviewing process through meaningful insights. By leveraging a novel customer-transaction bipartite graph through GNNs, we are able to obtain representations that characterize each entity given its surrounding context, and that can be used as a reference point of expected behavior used to score the anomaly of new transactions entering the system. We've shown that these representations also provide a unified entry point to build upon for other useful insights for the reviewing process, such as clustering the transactions of each customer, or identifying periods of abnormal activity of a customer under review. The goal is to incorporate this proposal within a broader system for AML reviewing with tailor-made visualizations that digest the provided insights and display them in an easy-to-understand manner, decreasing the burden and increasing the efficiency of AML analysts.

We evaluate our approach on a real-world banking dataset alongside several popular baselines, namely an MLP and LightGBM that inform their predictions exclusively through the raw feature information, and several graph-based variants that also exploit the structural information in the graph. We show that leveraging the information present in the underlying graph consistently improves performance, with the best method achieving an AUC of $\sim 95\%$ and AP of $\sim 96\%$, an improvement of $12.2$ and $6.2$ p.p. over the best non-graph baseline, respectively. We also show that, for the self-supervised objective of edge prediction, jointly training the encoder and decoder achieves superior results compared to pre-training the encoder on a separate self-supervised objective. Nevertheless, there is still room for exploration on how different self-supervised objectives can be combined to derive maximally informative representations.

A number of directions can be considered for future work, such as incorporating additional information in the form of different types of nodes e.g., merchants and card transactions. A particularly interesting direction of research would be to further exploit the temporal component present in the data through a sequential model that connects different graph snapshots in time. This would allow the representations of customers to capture the intrinsically evolving nature of a customer's transactional behavior, deriving representations aware of the past behavior not explicit in the input graph.

\begin{acks} 
The project CAMELOT (reference POCI-01-0247-FEDER-045915) leading to this work is co-financed by the ERDF - European Regional Development Fund through the Operational Program for Competitiveness and Internationalisation - COMPETE 2020, the North Portugal Regional Operational Program - NORTE 2020 and by the Portuguese Foundation for Science and Technology - FCT under the CMU Portugal international partnership.
\end{acks}

\printbibliography

\end{document}